\pdfoutput=1

\documentclass[11pt]{article}

\usepackage[final]{acl}

\usepackage{times}
\usepackage{latexsym}

\usepackage[T1]{fontenc}

\usepackage[utf8]{inputenc}

\usepackage{microtype}

\usepackage{inconsolata}

\usepackage{graphicx}
\usepackage{enumitem}
\usepackage{booktabs}
\usepackage{colortbl}

\definecolor{mycolor_green}{HTML}{D5E8D4}
\definecolor{mycolor_orange}{HTML}{FFE6CC}
\definecolor{mycolor_blue}{HTML}{DAE8FC}
\definecolor{mycolor_red}{HTML}{F8CECC}
\definecolor{case_green}{HTML}{B9E0A5}
\definecolor{case_yellow}{HTML}{FFE599}
\definecolor{case_blue}{HTML}{D4E1F5}

\title{From Personas to Talks: Revisiting the Impact of Personas on LLM-Synthesized Emotional Support Conversations}

\author{
 \textbf{Shenghan Wu\textsuperscript{1}}\quad
 \textbf{Yimo Zhu\textsuperscript{1}}\quad
 \textbf{Wynne Hsu\textsuperscript{1}}\quad
 \textbf{Mong-Li Lee\textsuperscript{1}}\quad
 \textbf{Yang Deng\textsuperscript{2}}
 \\
 \textsuperscript{1}National University of Singapore\quad
 \textsuperscript{2}Singapore Management University
 \\
 shenghan@u.nus.edu
}

\begin{document}
\maketitle
\begin{abstract}

The rapid advancement of Large Language Models (LLMs) has revolutionized the generation of emotional support conversations (ESC), offering scalable solutions with reduced costs and enhanced data privacy. This paper explores the role of personas in the creation of ESC by LLMs.
Our research utilizes established psychological frameworks to measure and infuse persona traits into LLMs, which then generate dialogues in the emotional support scenario. We conduct extensive evaluations to understand the stability of persona traits in dialogues, examining shifts in traits post-generation and their impact on dialogue quality and strategy distribution.
Experimental results reveal several notable findings: 1) LLMs can infer core persona traits, 2) subtle shifts in emotionality and extraversion occur, influencing the dialogue dynamics, and 3) the application of persona traits modifies the distribution of emotional support strategies, enhancing the relevance and empathetic quality of the responses.
These findings highlight the potential of persona-driven LLMs in crafting more personalized, empathetic, and effective emotional support dialogues, which has significant implications for the future design of AI-driven emotional support systems.

\end{abstract}

\section{Introduction}

In emotional support conversations (ESC), the supporter aims to help the seeker to reduce stress, overcome emotional issues, and promote mental well-being. Traditionally, ESC corpora have been developed through skilled crowdsourcing \cite{empatheticdialogue, esconv}, transcription of therapist sessions \cite{liu2023chatcounselor, shen-etal-2020-counseling}, or by compiling emotional question-answer pairs from online platforms \cite{sharma2020empathy, sun-etal-2021-psyqa, garg-etal-2022-cams, lahnala-etal-2021-exploring}. However, beyond the high costs, recent research \cite{sigir24-pp} highlights several limitations in these methods, including privacy concerns, variability in data quality, and fabricated user needs created by crowdworkers. With the advent of large language models (LLMs), their powerful generalization abilities enable high-quality data annotation and generation based on specific instructions \cite{llm-data-survey,acl24findings-llm-data}. Consequently, more and more recent studies \cite{acl23findings-augesc,zheng2024self,llm-esc,wu-etal-2024-ehdchat} investigate the use of LLMs to generate large-scale emotional support dialogue datasets across various scenarios via role-playing at lower costs. These efforts have significantly expanded the available ESC corpora.

\begin{figure}
    \centering
    \includegraphics[width=0.48\textwidth]{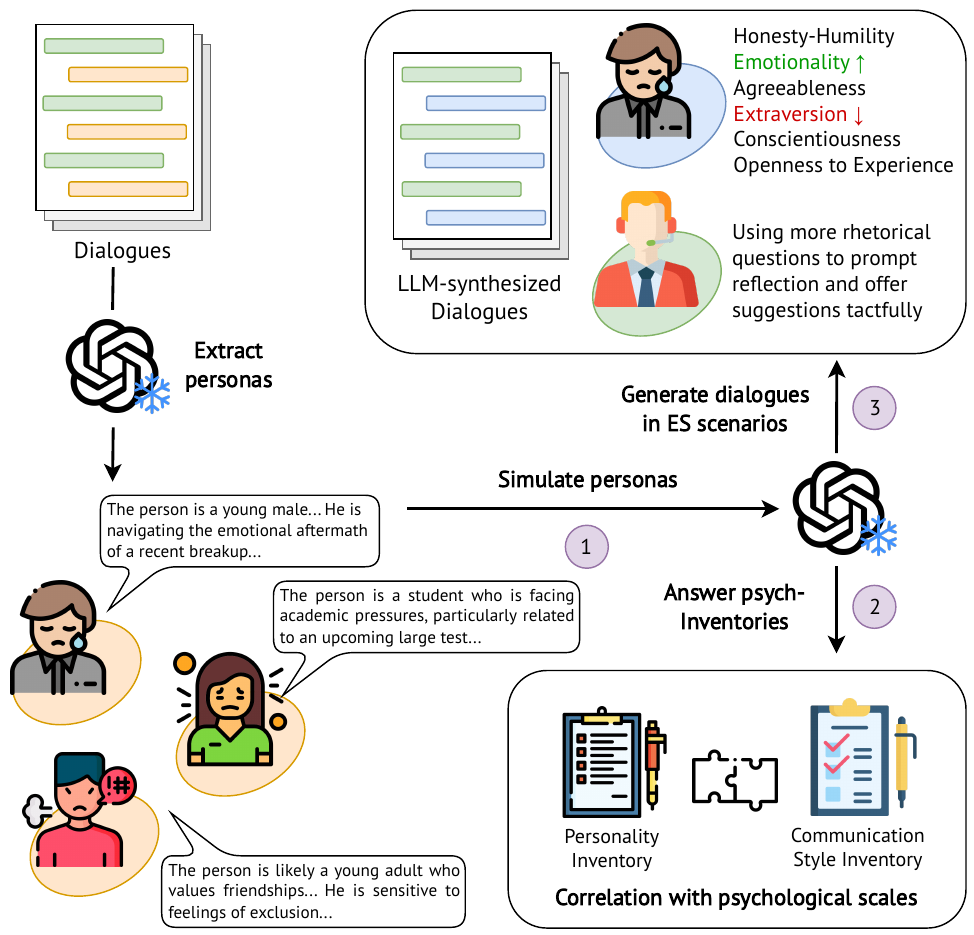}
    \caption{Overview of evaluating the impact of personas on LLM-synthesized emotional support dialogues.}
    \label{fig:main-pic}
\end{figure}

Despite the significant strides of LLMs in synthesizing ESC, a significant issue in generative AI annotations is the \textit{Lack of Human Intuition} \cite{sigir24-pp}.
Recent studies have demonstrated that effective emotional support requires careful consideration of individual differences, including personality traits, emotional states, and contextual factors \citep{ma2024personality, ait2023power, hernandez2023affective}. This understanding has led to increased research attention on the role of persona in emotional support dialogues \citep{zhao2024esc, cheng2022pal}, particularly as AI systems become more prevalent in providing such support. The importance of incorporating psychological perspectives in developing empathetic AI assistants has been emphasized by researchers. \citet{huang2023humanity} argues that psychological analysis of LLMs is crucial for creating more human-like and engaging interactions.
While recent work has made progress in measuring LLMs' personality characteristics using established psychological inventories \citep{frisch2024llm, safdari2023personality}, there remains a gap in understanding how these persona-related aspects influence the generation of emotional support dialogues. 

In this work, we aim to address this gap by investigating the relationship between LLM-generated emotional support dialogues and persona traits with psychological measurement.
Specifically, we propose a LLM-based simulation framework to answer the following critical research questions:
\begin{itemize}[leftmargin=*,nosep]
    \item \textbf{RQ1}: \textit{Can LLMs infer stable traits from persona in the emotional support scenario?}
    \item \textbf{RQ2}: \textit{Can dialogues generated by LLMs retain the original persona traits?}
    \item \textbf{RQ3}: \textit{How will the injected persona affect the LLM-simulated emotional support dialogues?}
\end{itemize}

As illustrated in Figure \ref{fig:main-pic}, we first extract personas from the existing datasets. Then we assess the capacity of LLMs to deduce stable traits from these personas. We further conduct a comparison between the personas derived from synthesized dialogues and the original personas to evaluate their stability during dialogue synthesis. Lastly, we investigate how these personas influence emotional support strategies.
As a result, we offer insights into the potential application of persona-driven dialogue synthesis in emotional support conversations. The key findings are summarized as follows:
\begin{itemize}[leftmargin=*]
    \item \textbf{LLMs can infer stable traits from personas like personalities and communication styles in emotional support scenarios.} We utilize LLMs to infer traits from personas, revealing a strong correlation between these traits.
    \item \textbf{LLM-simulated seekers tend to exhibit more emotionality and lower extraversion compared to their original personas.} After generating emotional support dialogues based on personas and extracting personas from these dialogues, there are slight shifts in persona traits.
    \item \textbf{Infusing persona traits into the generation of emotional support dialogues alters the distribution of strategies.} The LLM-simulated supporter tends to focus more on deeply understanding the seeker's problems and gently offers reassurance and encouragement.
\end{itemize}

\section{Related Works}


\subsection{Psychological Inventories}
Personality inventories are widely used in psychology to understand individuals, which predict distinctive patterns of interpersonal interaction across contexts.
These assessments are often structured, theory-driven, and standardized.
Prominent instruments include the Myers-Briggs Type Indicator (MBTI) \citep{mbti}, NEO-PI-R \citep{costa2008revised}, and Comrey Personality Scales (CPS) \citep{cps}. Among these, the HEXACO model \citep{ashton2009hexaco} is particularly notable, offering a framework that encompasses six factors: \textit{Honesty-Humility, Emotionality, Extraversion, Agreeableness, Conscientiousness, and Openness to Experience}.
\citet{norton1978foundation} introduces the foundational construct of a communicator style. \citet{de2013communication} further explored communication style as a six-dimensional model.
Extensive research \citep{capraro2002myers, costa1992four, lee2004psychometric} has demonstrated the reliability and validity of these inventories.

\subsection{Emotional Support Dialogues}
Early efforts on emotional support dialogues focused on collecting emotional question-answer data from online platforms \citep{medeiros2018using, sharma2020computational, turcan-mckeown-2019-dreaddit, garg-etal-2022-cams}. These datasets laid the groundwork for understanding user emotions, but were limited to single-turn interactions.
Empathetic Dialogue dataset \citep{empatheticdialogue} addressed this by introducing multi-turn dialogues, crowd-sourced to simulate diverse empathetic interactions.
ESConv \citep{esconv} further advanced the field by introducing emotional support strategies collected from psychological theories, enabling chatbots to use these strategies for more empathetic and contextually appropriate responses.
Subsequent studies proposed using graph networks to capture global emotions causes and user intentions \citep{peng2022control,acl23-esc}, combining multiple emotional support strategies to enhance empathy \citep{tu-etal-2022-misc}, developing proactive dialogue systems to lead the conversation towards positive emotions \cite{sigirap-tutorial,iclr24-ppdpp,tois-survey}, and implementing emotional support strategies and scenarios using LLMs to create the ExTES dataset \citep{zheng2024self}.

\subsection{Persona-Driven Emotional Support}
Recent advances have integrated personas into emotional support dialogues to enhance personalization and diversity. The ESC dataset \citep{zhang2024escot} introduced personas into the dialogue generation process. \citet{zhao2024esc} proposed a framework to extract personas from existing datasets for evaluation. Additionally, personas have been incorporated into chatbots to generate personalized responses \citep{tu2023characterchat, ait2023power, ma2024personality}.
These developments inspire our analysis of the relationship between personas, emotional support strategies, and dialogues, focusing on the extraction and use of personas in ESC.

\begin{figure}
    \centering
    \includegraphics[width=1\linewidth]{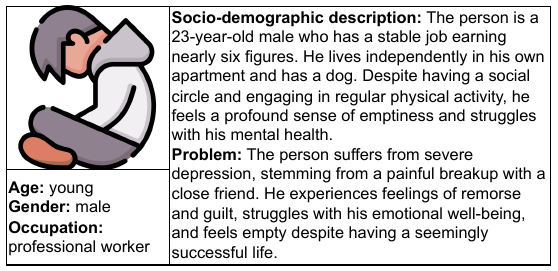}
    \caption{An example of persona card.}
    \label{fig:card-img}
\end{figure}

\section{Dataset Collection}
\label{sec:extract-persona}
In order to study the relationship between LLM-generated emotional support conversations (ESC) and persona traits, we first need to collect a set of non-synthetic ESC data with user personas as reference.
Specifically, we 
select three existing emotional support datasets, including
ESConv \citep{esconv}, CAMS \citep{garg-etal-2022-cams}, and Dreaddit \citep{turcan-mckeown-2019-dreaddit}. ESConv is a multi-turn emotional support dialogue dataset, while CAMS and Dreaddit are derived from Reddit posts discussing mental health issues.

\begin{table}
    \centering
    \small
    \begin{tabular}{l|ccc}\toprule
        \textbf{Dataset} & \textbf{ESConv} & \textbf{CAMS} & \textbf{Dreaddit} \\\midrule
        \textbf{Data Type} & dialogue & QA & QA \\
        \textbf{Num. of personas} & 1,155 & 1,140 & 730 \\
        \textbf{Avg. words of desc.} & 57.38 & 66.42 & 56.68 \\
        \textbf{Avg. words of prob.} & 33.92 & 32.02 & 27.75 \\
        \textbf{Num. of age} & 901 & 1,014 & 459 \\
        \textbf{Num. of gender} & 417 & 401 & 300 \\
        \textbf{Num. of occupation} & 926 & 968 & 542 \\\bottomrule   
    \end{tabular}
    \caption{Statistics of the extracted persona cards.}
    \label{tab:stat-analysis}
\end{table}

To obtain the user persona traits from these ESC data, we extract \textbf{\textit{age, gender, occupation, socio-demographic description, and problem 
}} from the datasets using gpt-4o-mini.
The prompt utilized for extracting these basic persona is provided 
in Appendix \ref{sec:persona-card-prompt}.
Since the supporters usually focus on the seeker's emotions and their main tasks are to provide emotional support, the supporters' responses in the datasets lack personal information.
Due to the limited information available about supporters, we only extract seeker personas from these datasets.
After extraction, we prompt LLMs
(see Appendix \ref{sec:persona-card-prompt})
to filter the personas to ensure they include the individual's emotions and the events they are experiencing, along with a clear socio-demographic background that provides a comprehensive sense of their identity. Ultimately, we obtain 
1,155 personas from ESConv,
1,140 personas from CAMS, and 730 personas from Dreaddit. An example of a basic persona card is illustrated in Figure \ref{fig:card-img}. 
The detailed statistics of datasets to be studied are summarized in Table \ref{tab:stat-analysis}.

\begin{figure}
    \centering
    \includegraphics[width=0.4\textwidth]{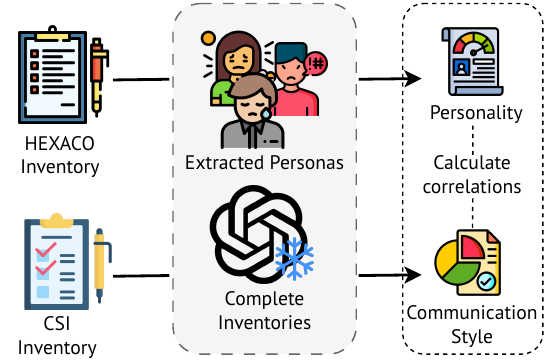}
    \caption{Diagram of the process for the measurement of persona traits.}
    \label{fig:rq1}
\end{figure}

\section{Measurement of Persona Traits (RQ1)}
\label{sec:measure-traits}
Personality and communication style play crucial roles in emotional support conversations. Personality influences how a seeker may emotionally respond to various scenarios \citep{hughes2020personality, komulainen2014effect}, while communication style reflects the way someone conveys their thoughts and emotions during the interaction \citep{van2023effects}. These factors significantly impact the dynamics and outcomes of emotional support. Previous studies have demonstrated the effectiveness of personas in guiding the responses of emotional support systems \citep{cheng2022pal, han2024persona}. In this section, we investigate whether the personality and communication style inferred by LLMs based on persona cards are correlated. In other words, we examine whether LLMs can infer stable traits from persona cards in emotional support conversations.

\begin{table}[t]
        \begin{minipage}{0.48\textwidth}
        \centering
        \small
        \begin{tabular}{c|cccccc} \toprule
                   & Expr. & Prec. & Verb. & Ques. & Emot. & Impr. \\\midrule
             Extr. & \cellcolor{mycolor_green}{\textbf{.54}} & .15 & -.21 & .36 & -.39 & .04 \\
             Cons. & .34 & \cellcolor{mycolor_green}{\textbf{.34}} & -.11 & .16 & -.36 & .02 \\
             Agre. & .21 & .15 & \cellcolor{mycolor_green}{\textbf{-.39}} & .12 & -.19 & -.05 \\
             Open. & .41 & .25 & -.23 & \cellcolor{mycolor_green}{\textbf{.47}} & -.09 & .09 \\
             Emot. & -.32 & -.11 & -.04 & -.21 & \cellcolor{mycolor_green}{\textbf{.45}} & -.10 \\
             Hone. & -.24 & -.01 & -.17 & -.27 & .05 & \cellcolor{mycolor_green}{\textbf{-.18}} \\\bottomrule
            \end{tabular}
            \caption{Correlations between CSI and HEXACO from ESConv measured by \textit{gpt-4o-mini}.
            P\_value of all metrics are less than 0.01.}
            \label{tab:ESConv-seeker}
    \vspace{3mm}
\end{minipage}
\begin{minipage}{0.48\textwidth}
        \centering
        \small
        \begin{tabular}{c|cccccc} \toprule
                   & Expr. & Prec. & Verb. & Ques. & Emot. & Impr. \\\midrule
             Extr. & \cellcolor{mycolor_green}{\textbf{.63}} & .01 & -.21 & \cellcolor{mycolor_red}{\textbf{.50}} & .07 & -.11 \\
             Cons. & .15 & \cellcolor{mycolor_green}{\textbf{.48}} & -.22 & .04 & -.18 & -.19 \\
             Agre. & .13 & .14 & \cellcolor{mycolor_green}{\textbf{-.59}} & -.02 & .06 & \cellcolor{mycolor_red}{\textbf{-.39}} \\
             Open. & .39 & .01 & -.28 & \cellcolor{mycolor_green}{.46} & .21 & -.11 \\
             Emot. & -.24 & -.18 & .00 & -.22 & \cellcolor{mycolor_green}{\textbf{.32}} & -.06 \\
             Hone. & .06 & .28 & -.42 & .00 & .05 & \cellcolor{mycolor_green}{-.37} \\\bottomrule
            \end{tabular}
            \caption{Correlations between CSI and HEXACO from ESConv measured by \textit{Claude-3.5-haiku}.}
            \label{tab:esconv-seeker-claude}
    \vspace{3mm}
\end{minipage}
\begin{minipage}{0.48\textwidth}
        \centering
        \small
        \begin{tabular}{c|cccccc} \toprule
                   & Expr. & Prec. & Verb. & Ques. & Emot. & Impr. \\\midrule
             Extr. & \cellcolor{mycolor_green}{\textbf{.28}} & .26 & \cellcolor{mycolor_red}{\textbf{-.21}} & .15 & -.33 & -.11 \\
             Cons. & .13 & \cellcolor{mycolor_green}{\textbf{.55}} & -.16 & -.01 & -.42 & -.04 \\
             Agre. & -.02 & -.13 & \cellcolor{mycolor_green}{-.19} & .03 & .10 & -.05 \\
             Open. & .07 & -.12 & -.15 & \cellcolor{mycolor_green}{\textbf{.32}} & .08 & -.10 \\
             Emot. & -.18 & -.32 & 0.06 & -.02 & \cellcolor{mycolor_green}{\textbf{.48}} & .01 \\
             Hone. & -.08 & -.11 & -.20 & .06 & .07 & \cellcolor{mycolor_green}{\textbf{-.13}} \\\bottomrule
            \end{tabular}
            \caption{Correlations between CSI and HEXACO from ESConv measured by \textit{LLaMA-3.1-8B-Instruct}.}
            \label{tab:esconv-seeker-llama}
    \end{minipage}
\end{table}

\subsection{Experimental Setups}
\label{sec:expand}
Previous studies showed that LLMs have sufficient capability to capture certain level of human traits from dialogues \citep{jiang2024personallm, porvatov2024big}.
To further investigate whether LLMs can infer stable traits, we utilize HEXACO model \citep{lee2004psychometric} to assess personality and Communication Styles Inventory (CSI) \citep{de2013communication} to evaluate communication styles.
The HEXACO model is a personality framework consisting of six dimensions: \textit{Honesty-Humility} (H), \textit{Emotionality} (E), \textit{Extraversion} (X), \textit{Agreeableness} (A), \textit{Conscientiousness} (C), and \textit{Openness to Experience} (O). For our study, we use the HEXACO-60 inventory \citep{ashton2009hexaco} to assess the personalities represented in the persona cards.
The CSI identifies six domain-level communicative behavior scales: \textit{Expressiveness}, \textit{Preciseness}, \textit{Verbal Aggressiveness}, \textit{Questioningness}, \textit{Emotionality}, and \textit{Impression Manipulativeness}.
Each dimension of the HEXACO personality model has the strongest correlation with a specific communication style dimension in the CSI. The correlations between these dimensions are introduced as follows
(left - HEXACO, right - CSI) \citep{de2013communication}:
\begin{itemize}[leftmargin=*,nosep]
    \item \textit{Extraversion <-> Expressiveness}
    \item \textit{Conscientiousness <-> Preciseness}
    \item \textit{Agreeableness <-> Verbal Aggressiveness}
    \item \textit{Openness to Experience <-> Questioningness}
    \item \textit{Emotionality <-> Emotionality}
    \item \textit{Honesty-Humility <-> Impression Manipulativeness}
\end{itemize}

If LLMs can infer stable personality and communication style based on emotional support dialogues, the strongest correlations in the personality scores and communication style scores obtained from completing the inventories will align with those shown above.

Following previous research \citep{ji2024persona}, we prompt the LLM to generate descriptions for each dimension based on the extracted socio-demographic information, incorporating these descriptions into persona cards. Then, we prompt LLMs to predict answers to the HEXACO and CSI inventories using the persona card.
These prompts are presented in Appendix \ref{sec:persona-card-prompt}.
We run our experiments on multiple open-source (LLaMA-3.1-8B-Instruct\footnote{meta-llama/Llama-3.1-8B-Instruct}) and close-source LLMs (GPT-4o-mini\footnote{gpt-4o-mini-2024-07-18} and Claude-3.5-Haiku\footnote{claude-3-5-haiku-20241022}), and we set temperature as 0 to get stable results. 

\subsection{Results and Discussions}
Based on the responses from these inventories, we calculate the HEXACO and CSI dimension scores for each dataset. To evaluate whether LLMs can infer stable traits from persona cards in an emotional support context, we then compute the correlations between the HEXACO and CSI dimensions within each dataset.
Each scale provides a score corresponding to each response, along with the dimension to which each question belongs. Based on the LLM's responses, we calculate the scores for each dimension. Finally, we use Pearson correlation to analyze the relationships between each dimension of HEXACO and CSI.
Tables \ref{tab:ESConv-seeker}, \ref{tab:esconv-seeker-claude}, and \ref{tab:esconv-seeker-llama} present the correlations between HEXACO and CSI dimensions in the ESConv dataset \citep{esconv} as measured by three different LLMs. Similarly, Tables \ref{tab:gpt-CAMS}, \ref{tab:cams-seeker-claude}, and \ref{tab:cams-seeker-llama} show the correlations in the CAMS dataset \citep{garg-etal-2022-cams}, while Tables \ref{tab:gpt-Dreaddit}, \ref{tab:dreaddit-seeker-claude}, and \ref{tab:dreaddit-seeker-llama} provide the correlations observed in the Dreaddit dataset \citep{turcan-mckeown-2019-dreaddit}.

Experimental results demonstrate that on all three test datasets, GPT-4o-mini exhibits the strongest correlations between the six dimensions of HEXACO and CSI, aligning well with findings from the established psychological theory. For instance, extraversion from HEXACO model shows the strongest correlation with expressiveness from CSI model. This indicates that GPT-4o-mini can reliably infer persona traits relevant to emotional support dialogues from persona cards. However, we observed some discrepancies in the correlation analyses for LLaMA-3.1-8B-Instruct and Claude-3.5-Haiku. Specifically, LLaMA-3.1-8B-Instruct incorrectly associates verbal aggressiveness with extraversion and conscientiousness, suggesting that higher verbal aggressiveness implies greater extraversion in seekers. Meanwhile, Claude-3.5-Haiku incorrectly links questioningness with conscientiousness, implying that seekers who ask more questions are more extraverted. These inconsistencies highlight potential limitations in the ability of these models to interpret certain persona traits accurately. Overall, our findings suggests that \textbf{LLMs are capable of inferring stable persona traits from personas in emotional support scenarios}, though some inconsistencies exist.

\section{Persona Consistency in LLM-simulated Emotional Support Dialogues (RQ2)}

After demonstrating that LLMs can reliably infer stable traits from personas, we further investigate whether these persona traits remain consistent during the LLM-based dialogue generation process.

\begin{figure}[t]
        \centering
        \includegraphics[width=0.48\textwidth]{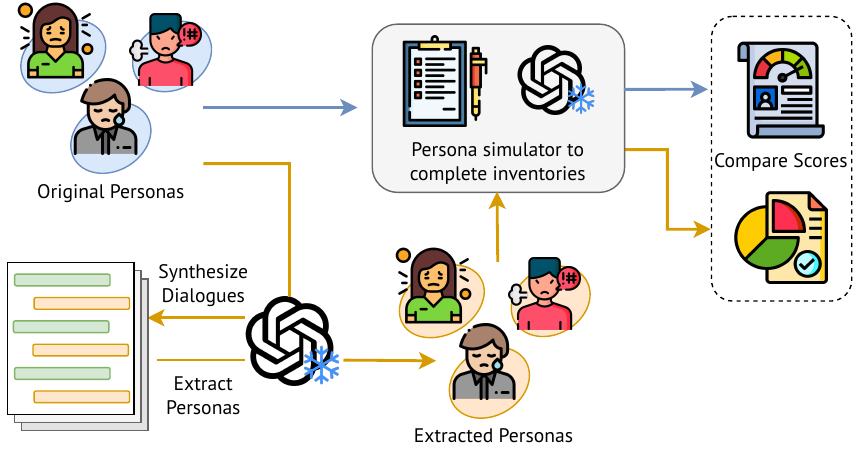}
        \caption{Diagram of the process for studying the persona consistency in LLM-simulated ESC.}
        \label{fig:rq2}
\end{figure}

\begin{figure}[t]
    \centering
    \begin{minipage}{0.48\textwidth}
        \centering
        \includegraphics[width=\textwidth]{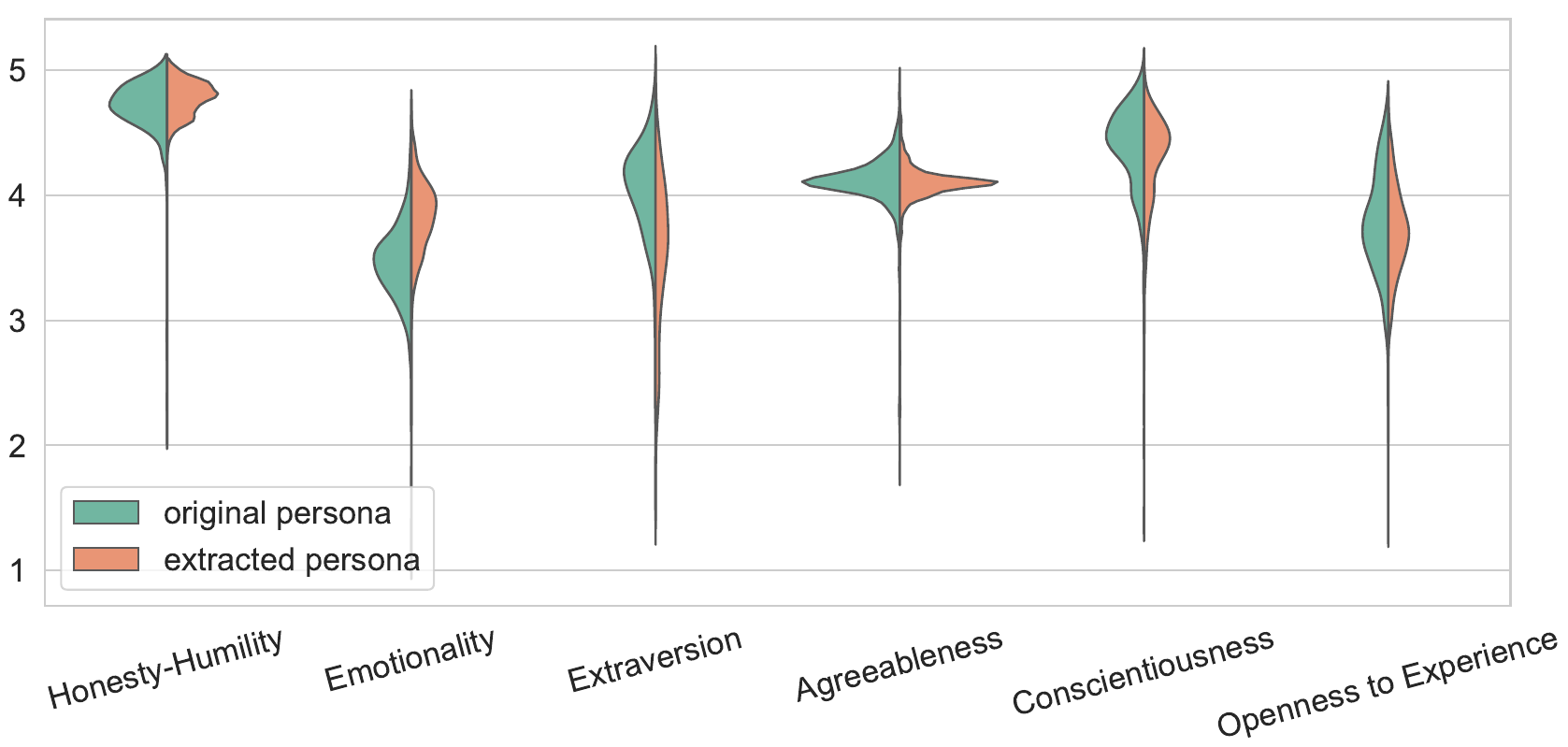}
        \caption{Comparison of HEXACO scores between the original persona and the one extracted from the dialogue generated by \textit{gpt-4o-mini}.}
        \label{fig:hexaco-violin}
        \vspace{0.2cm}
    \end{minipage}
    \begin{minipage}{0.48\textwidth}
        \centering
        \includegraphics[width=\textwidth]{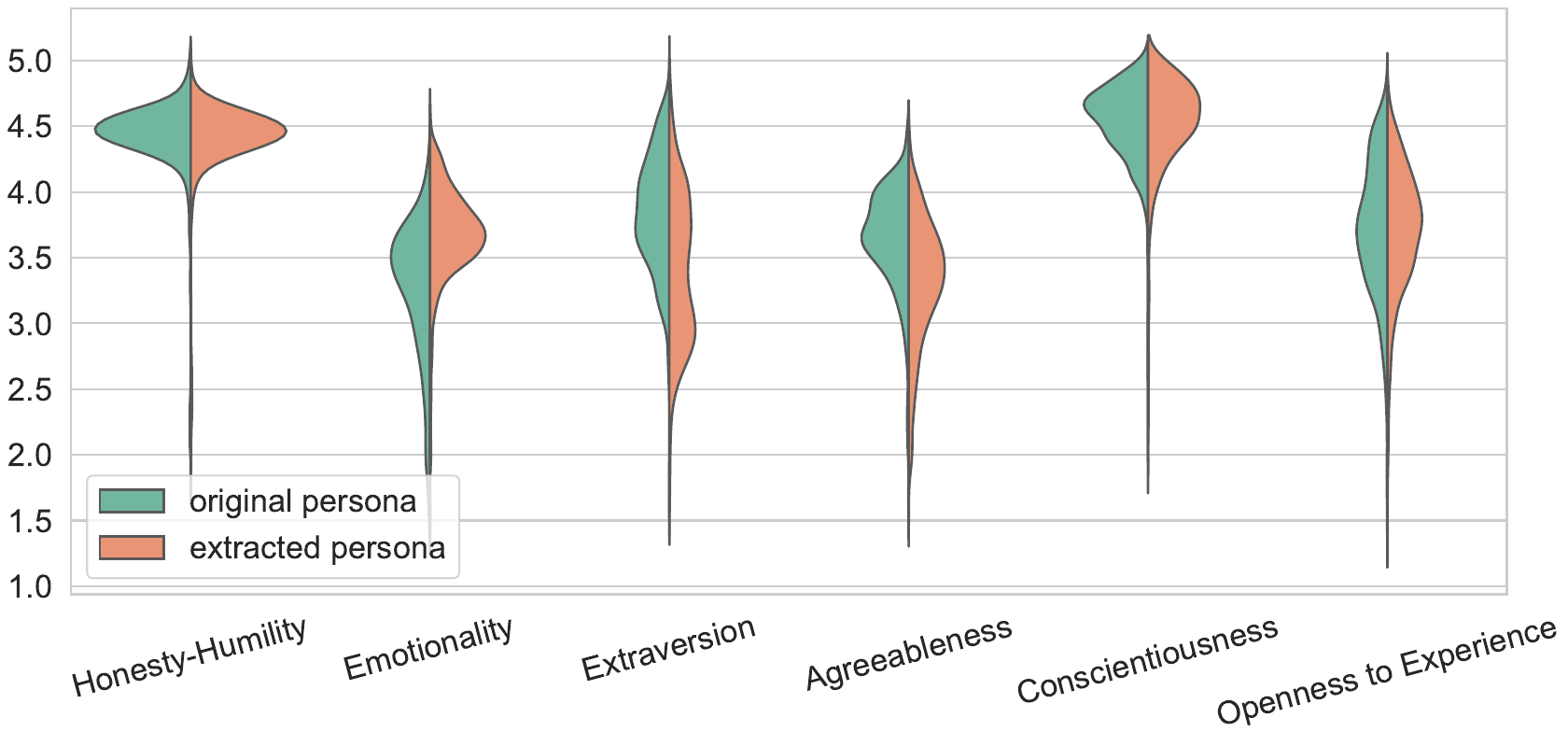}
        \caption{Comparison of HEXACO scores between the original persona and the one extracted from the dialogue generated by \textit{claude-3.5-haiku}.}
        \label{fig:hexaco-violin-claude}
        \vspace{0.2cm}
    \end{minipage}
    \begin{minipage}{0.48\textwidth}
        \centering
        \includegraphics[width=\textwidth]{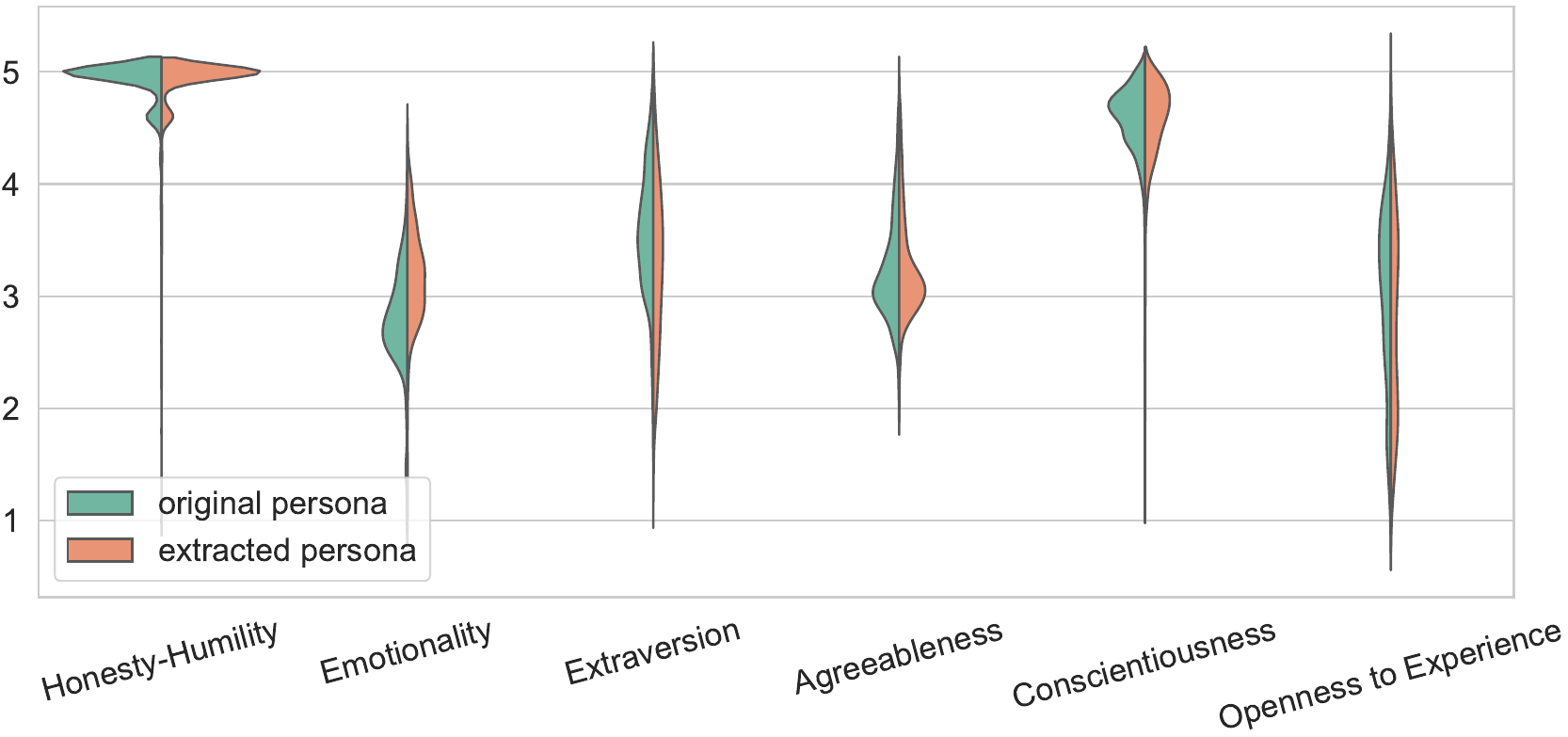}
        \caption{Comparison of HEXACO scores between the original persona and the one extracted from the dialogue generated by \textit{LLaMA-3.1-8B-Instruct}.}
        \label{fig:hexaco-violin-llama}
    \end{minipage}
\vspace{-0.2cm}
\end{figure}

\subsection{Experimental Setups}
To evaluate whether LLMs maintain consistent persona traits after dialogue generation, we conducted an extensive analysis using 1,000 randomly selected personas from PersonaHub \citep{chan2024scaling}, ensuring a diverse range of characteristics. These personas initially consist of simple descriptions, such as "\textit{A fearless and highly trained athlete who can perform complex and dangerous physical sequences}", that we systematically enhanced through LLM-based expansion. The enhancement process involved adding socio-demographic details (age, gender, occupation) and specific trait-indicative statements aligned with HEXACO and CSI dimensions. We then quantified these enhanced personas by generating HEXACO and CSI dimension scores using the methodology described in Section \ref{sec:expand}.
As LLMs can be effectively shaped to emulate human-beings \citep{frisch2024llm, wang2023rolellm}, we use these enriched personas to generate 
emotional support dialogues where each persona acted as a seeker in contextually relevant scenarios. For instance, an athlete discussing an injury-related emotional challenge. Following dialogue generation, we applied the extraction method outlined in Section \ref{sec:extract-persona} to derive persona characteristics from the generated conversations. We then calculated HEXACO and CSI scores from these extracted persona. By comparing these extracted scores with the original scores assigned to the input personas, we could assess the consistency of trait representation after the dialogue generation process. The complete set of prompts used for dialogue generation and trait extraction is provided in Appendix \ref{sec:prompt-stability}.

\begin{figure}
    \centering
    \includegraphics[width=0.4\textwidth]{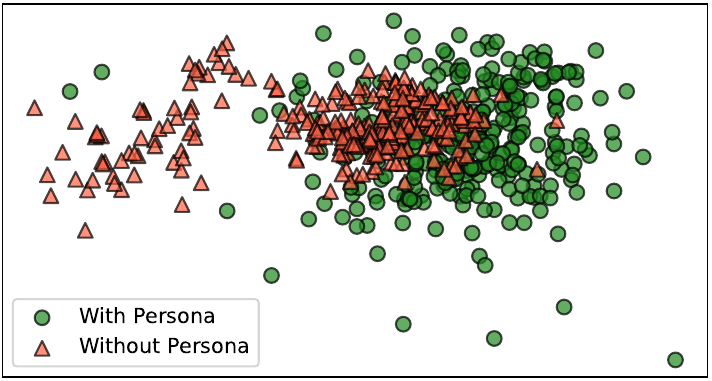}
    \caption{Distribution of personality scores, reduced to 2D, obtained from dialogues w/o persona injection.}
    \label{fig:compare}
\end{figure}

\subsection{Results and Discussions}
\label{sec:rq2-results}
The comparisons of scores of persona traits are shown in the Figure \ref{fig:hexaco-violin}, \ref{fig:hexaco-violin-claude}, and \ref{fig:hexaco-violin-llama}.
The results show that the original personas and the personas extracted from the generated dialogues share similar distributions across four personality traits: Honesty-Humility, Agreeableness, Conscientiousness, and Openness to Experience.
This indicates that the personas maintain consistent traits in the synthetic emotional support dialogues.
However, we notice that the extracted personas tend to have higher Emotionality and lower Extraversion compared to the original personas. We believe this may be because the seeker in the emotional support dialogues is dealing with emotional issues, making them more emotional and less outgoing.
A similar pattern is observed in the comparison of CSI scores (see Table \ref{fig:csi-violin}, \ref{fig:csi-violin-claude}, and \ref{fig:csi-violin-llama} in Appendix \ref{sec:persona-consistency}). 
This indicates that \textbf{the persona traits generally maintain consistent in the synthetic emotional support dialogues with only slight shifts}.

\begin{figure}[t]
        \centering
        \includegraphics[width=0.48\textwidth]{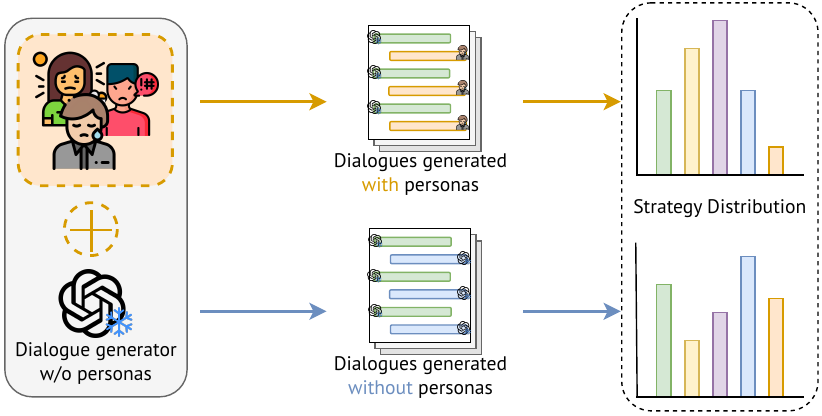}
        \caption{Diagram of the process for studying the impact of persona on LLM-simulated ESC.}
        \label{fig:rq3}
\end{figure}

\subsection{Ablation Study}
Prior research \citep{huang2023revisiting, pan2023llms, frisch2024llm} suggests that LLMs possess inherent personality traits. To investigate how these intrinsic traits might influence emotional support dialogue generation, we conducted a comparative analysis between dialogues generated with and without predefined personas.
We extracted the implicit personas manifested in the generated conversations, calculated their corresponding personality scores.
In Figure \ref{fig:compare}, we used PCA to project these scores into a 2D space for visualization. The resulting distribution reveals a marked difference between the two conditions: dialogues generated without persona injection exhibit a more concentrated distribution, whereas those with predefined personas cover a broader range of personality traits.
This contrast suggests that externally provided personas influence the personality traits manifested in dialogue generation, persona injection can guide and shape the dialogue generation process.

\section{Impact of Persona on LLM-simulated Emotional Support Dialogues (RQ3)}
Many efforts have been made on exploring how to incorporate emotional support strategies into emotional support dialogues. \citet{zheng2024self} use LLMs to introduce emotional support strategies in synthetic dialogues. \citet{zhang2024escot} further involve the concept of persona into these dialogues and analyze the usage of strategies.
This raises an important question: 
does adding a persona have an impact on the way emotional support strategies (definitions of each strategy are shown in Appendix \ref{sec:definitions-strategies})
are distributed in dialogues? 
To investigate it, we use ESConv dialogues as history and instruct LLMs to predict how a new conversation might unfold after some time. Each future dialogue is generated in two versions: with and without persona traits (personality and communication style scores).
Since ESConv provides real-world context and continuations generated by LLMs can preserve the semantic distribution of dialogues \citep{fan2025consistency}, our approach ensures diversity and realism in emotional support scenarios.

\begin{table}[t]
     \begin{minipage}{0.48\textwidth}
    \centering
     \small
    \begin{tabular}{l|c|c} \toprule
        \textbf{Strategy} & \textbf{w/ PT} & \textbf{w/o PT} \\\midrule
        question & \cellcolor{mycolor_green}{27.23\%} & \cellcolor{mycolor_red}{16.45\%} \\
        restatement or paraph. & \cellcolor{mycolor_red}{3.61\%} & \cellcolor{mycolor_green}{10.57\%} \\
        reflection of feelings & \cellcolor{mycolor_green}{11.75\%} & \cellcolor{mycolor_red}{11.33\%} \\
        self-disclosure & \cellcolor{mycolor_red}{2.64\%} & \cellcolor{mycolor_green}{10.64\%} \\
        affirmation and reass. & \cellcolor{mycolor_green}{29.72\%} & \cellcolor{mycolor_red}{21.06\%} \\
        providing suggestions & \cellcolor{mycolor_green}{16.92\%} & \cellcolor{mycolor_red}{14.25\%} \\
        information & \cellcolor{mycolor_red}{0.78\%} & \cellcolor{mycolor_green}{4.85\%} \\
        others & \cellcolor{mycolor_red}{8.45\%} & \cellcolor{mycolor_green}{10.88\%} \\\bottomrule
    \end{tabular}
    \caption{Strategy distribution on two different groups of synthesized dialogues by \textit{gpt-4o-mini}, one generated with persona traits (PT), the other without.}
    \label{tab:strategies-distribution}
 \vspace{3mm}
\end{minipage}
\begin{minipage}{0.48\textwidth}
    \centering
     \small
    \begin{tabular}{l|c|c} \toprule
        \textbf{Strategy} & \textbf{w/ PT} & \textbf{w/o PT} \\\midrule
        question & \cellcolor{mycolor_green}{33.10\%} & \cellcolor{mycolor_red}{27.31\%} \\
        restatement or paraph. & \cellcolor{mycolor_red}{0.69\%} & \cellcolor{mycolor_green}{0.96\%} \\
        reflection of feelings & \cellcolor{mycolor_green}{21.19\%} & \cellcolor{mycolor_red}{18.15\%} \\
        self-disclosure & \cellcolor{mycolor_red}{2.22\%} & \cellcolor{mycolor_green}{13.97\%} \\
        affirmation and reass. & \cellcolor{mycolor_green}{19.32\%} & \cellcolor{mycolor_red}{17.41\%} \\
        providing suggestions & \cellcolor{mycolor_green}{19.01\%} & \cellcolor{mycolor_red}{15.41\%} \\
        information & \cellcolor{mycolor_red}{4.45\%} & \cellcolor{mycolor_green}{6.77\%} \\
        others & \cellcolor{mycolor_green}{0.02\%} & \cellcolor{mycolor_green}{0.02\%} \\\bottomrule
    \end{tabular}
    \caption{Strategy distribution on two different groups of synthesized dialogues by \textit{claude-haiku-3.5}.}
    \label{tab:strategies-distribution-claude}
 \vspace{3mm}
\end{minipage}
\begin{minipage}{0.48\textwidth}
    \centering
     \small
    \begin{tabular}{l|c|c} \toprule
        \textbf{Strategy} & \textbf{w/ PT} & \textbf{w/o PT} \\\midrule
        question & \cellcolor{mycolor_green}{12.70\%} & \cellcolor{mycolor_red}{12.34\%} \\
        restatement or paraph. & \cellcolor{mycolor_red}{6.51\%} & \cellcolor{mycolor_green}{7.56\%} \\
        reflection of feelings & \cellcolor{mycolor_green}{18.44\%} & \cellcolor{mycolor_red}{18.06\%} \\
        self-disclosure & \cellcolor{mycolor_red}{7.69\%} & \cellcolor{mycolor_green}{9.86\%} \\
        affirmation and reass. & \cellcolor{mycolor_green}{21.42\%} & \cellcolor{mycolor_red}{18.91\%} \\
        providing suggestions & \cellcolor{mycolor_red}{13.33\%} & \cellcolor{mycolor_green}{13.40\%} \\
        information & \cellcolor{mycolor_red}{2.48\%} & \cellcolor{mycolor_green}{4.25\%} \\
        others & \cellcolor{mycolor_green}{17.43\%} & \cellcolor{mycolor_red}{15.62\%} \\\bottomrule
    \end{tabular}
    \caption{Strategy distribution on two different groups of synthesized dialogues by \textit{LLaMA-3.1-8B-Instruct}.}
    \label{tab:strategies-distribution-llama}
    \end{minipage}
\end{table}

\begin{table}[t]
    \centering
    \small
    \begin{tabular}{l|c|c} \toprule
        \textbf{Strategy} & \textbf{HEXACO} & \textbf{CSI} \\\midrule
        question & 27.83\% & 27.23\% \\
        restatement or paraph. & 3.72\% & 3.61\% \\
        reflection of feelings & 12.48\% & 11.75\% \\
        self-disclosure & 3.41\% & 2.64\% \\
        affirmation and reass. & 28.96\% & 29.72\% \\
        providing suggestions & 16.44\% & 16.92\% \\
        information & 0.50\% & 0.78\% \\
        others & 6.66\% & 8.45\% \\\bottomrule
    \end{tabular}
    \caption{Strategy distribution on two different groups of synthesized dialogues, one generated with HEXACO scores, the other with CSI scores.}
    \label{tab:strategies-distribution-HEXACO-CSI}
\end{table}

\subsection{Analysis of Emotional Support Strategies}
As demonstrated in Tables \ref{tab:strategies-distribution} and \ref{tab:strategies-distribution-claude}, the distribution of emotional support strategies differs significantly between dialogues generated with and without persona traits for both \textit{gpt-4o-mini} and \textit{claude-3.5-haiku}. While Table \ref{tab:strategies-distribution-llama} shows that \textit{llama-3.1-8B-instruct} exhibits less pronounced differences in strategy distributions between persona and non-persona conditions, the trends remain consistent with those observed in other models.
The analysis revealed that supporters in persona-enhanced dialogues demonstrated a greater tendency to validate seekers' emotions through questioning and provided more effective emotional comfort, as evidenced by significantly higher rates of questioning, affirmation, and reassurance strategies compared to non-persona dialogues. Conversely, supporters in dialogues without persona traits emphasized problem explanation and relied more heavily on self-disclosure, 
while prior research \citep{meng2021emotional} indicate that self-disclosing chatbots perform poorly when emotional support is absent and the clear boundaries are not established.
The distributions of emotional support strategies generated using either personality scores or communication style scores, as shown in Table \ref{tab:strategies-distribution-HEXACO-CSI}, demonstrate remarkable similarity. This alignment can be attributed to the strong correlation between these trait measures, as discussed in Section \ref{sec:measure-traits}. These results provide compelling evidence that \textbf{persona traits significantly influence the distribution and application of emotional support strategies in dialogues}.

\begin{table}[t]
    \centering
    \small
    \begin{tabular}{r|ccccc}\toprule
    \textbf{w/ vs. w/o PT} & \textbf{Win} & \textbf{Tie} & \textbf{Loss} \\
    \midrule
    Suggestion & \textbf{38\%} & 27\% & 35\% \\
    Consistency & 27\% & \textbf{54\%} & 19\% \\
    Comforting & \textbf{38\%} & 28\% & 34\% \\
    Identification & \textbf{37\%} & 30\% & 33\% \\
    Overall & \textbf{39\%} & 27\% & 34\% \\
    \bottomrule
    \end{tabular}
    \caption{Human evaluation compares dialogues generated with and without personas. \textbf{Win} indicates that the dialogues generated with persona outperforms the one generated without persona on the given indicator.
    }
    \label{tab:human-compare}
\end{table}

\begin{figure*}[t]
    \centering
    \includegraphics[width=0.98\linewidth]{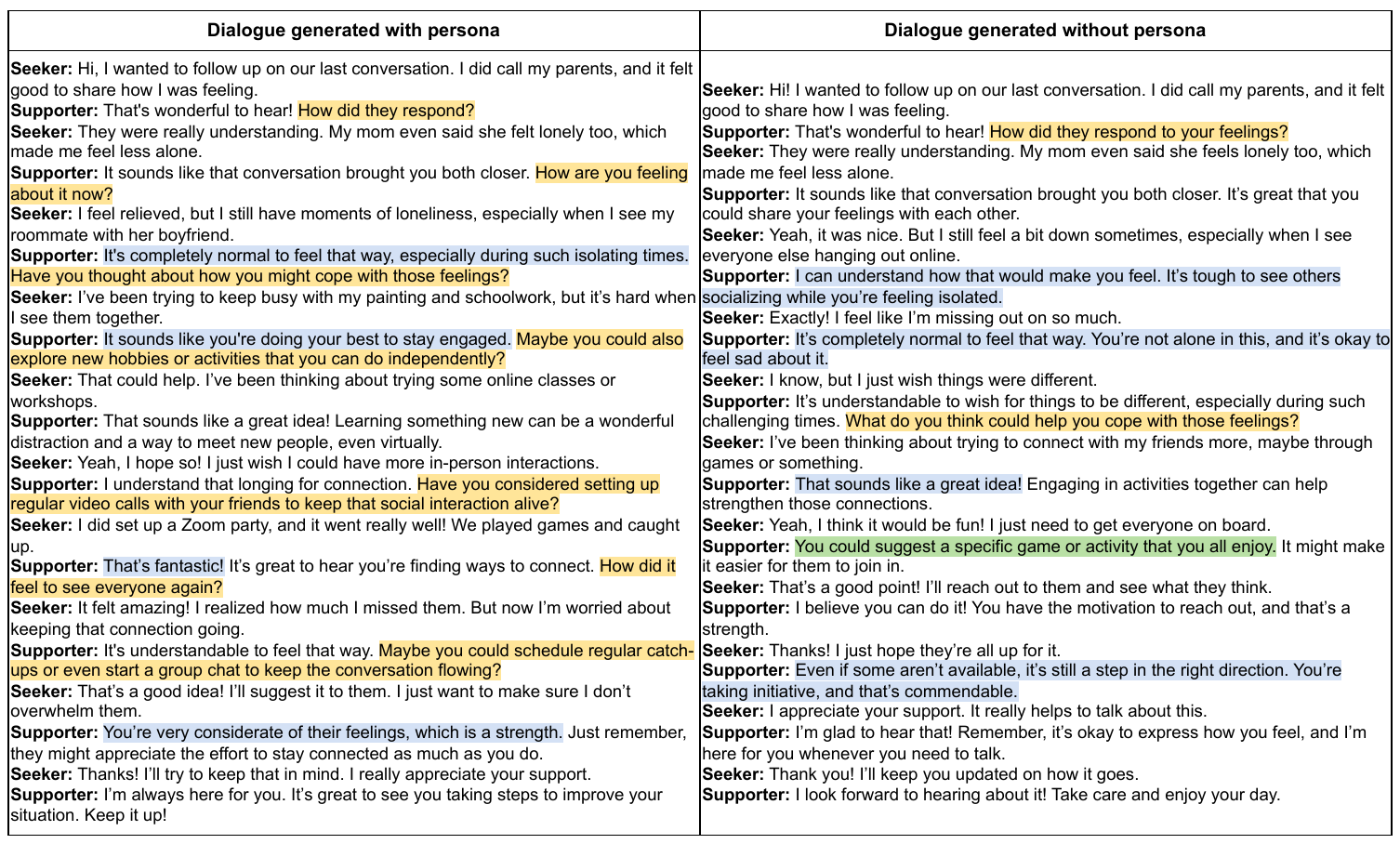}
    \caption{Case study. \colorbox{case_blue}{Blue} indicates that the supporter directly provides emotional support. \colorbox{case_green}{Green} signifies the supporter offers direct suggestions. \colorbox{case_yellow}{Yellow} means that the supporter provides suggestions through rhetorical questions or guides the seeker to reflect.}
    \label{fig:case-short}
\end{figure*}

\subsection{Human Evaluation} 
To intuitively reveal the impact of persona on LLM-simulated emotional support conversations, we conduct human evaluation by comparing the generated dialogues with and without persona traits, using the following indicators:
Annotators evaluate the dialogues based on the following metrics:
(1) \textbf{Suggestion} evaluates how effectively the supporter provided helpful advice.
(2) \textbf{Consistency} assesses whether participants consistently maintained their roles and exhibit coherent behavior.
(3) \textbf{Comforting} examines the supporter's ability to provide emotional support to the seeker.
(4) \textbf{Identification} determines which supporter delved deeper into the seeker's situation and was more effective in identifying issues.
(5) \textbf{Overall} assesses the overall performance about these two groups of dialogues.
We recruited 10 native English speakers from undergraduate students in various disciplines, who had previously completed various annotation tasks and were experienced in the field. Each evaluator was compensated 
at a rate of approximately 20 USD per hour
and was fully informed about the tasks they needed to complete.
They reviewed 50 randomly selected groups of instances from the generated dialogues with and without personas.

The result of human evaluation is shown in Table \ref{tab:human-compare}.
Except for the Consistency indicator, which showed similar performance for both dialogue groups, the dialogues generated with personas outperformed those without.
Based on dialogue strategies and observations, we believe that the dialogues generated with personas are better at using rhetorical questions to identify the seeker's issues and gently offer suggestions, making the conversations more in-depth and comforting.

\subsection{Case Study} 
A case study supports our point can be found in Figure \ref{fig:case-short}, and the persona card and dialogue history can be found in Appendix \ref{sec:case-study}.
Although we found that these dialogues provided the same level of direct emotional support, in the dialogue generated with persona, there is a tendency to use rhetorical questions to encourage the seeker to reflect and explore the content more deeply, while also offering suggestions more tactfully. In contrast, the dialogue generated without persona is more likely to provide direct affirmations or suggestions.
In psychology, \textit{when a message is highly relevant to the recipient and they are already motivated to process it, rhetorical questions can enhance the persuasiveness of messages with weak arguments} \citep{petty1981effects}. Since emotional support is often regarded as a form of weak argument \citep{petty2012communication}, incorporating rhetorical questions into a supporter's response can make it more acceptable to the seeker, and facilitate deeper and more meaningful conversations. This observation highlights the role of personas in enhancing the quality of dialogue generation.

\section{Conclusion}
This analytical study highlights the potential of incorporating personas into LLM-generated emotional support dialogues to enhance effectiveness and human-likeness. 
Our findings confirm that LLMs can infer stable traits from personas and maintain key persona characteristics while revealing shifts in emotionality and extraversion traits that affect dialogue. Personas not only improve the empathetic quality of responses but also influence the distribution of emotional support strategies, ensuring dialogues are more personalized and contextually appropriate. We encourage future researchers to build on our findings to develop more adaptive, effective, and robust ESC chatbots.

\section*{Acknowledgments}
This research was supported by the Singapore Ministry of Education (MOE) Academic Research Fund (AcRF) Tier 1 grant (No. MSS24C012).

\section*{Limitations}
The reliance on LLM outputs introduces potential biases inherent in the model's training data. These biases may have influenced both the extraction and simulation of personas, possibly affecting the accuracy of the results comparing to real persons.
Our approach follows the recent researches about extracting personas using LLMs \citep{ji2024persona, zhao2024esc}, which may still have limitations in accurately capturing and representing complex human traits.
Future researchers may need to investigate the impact of inherent biases in LLMs on persona extraction and dialogue simulation based on these personas.

Additionally, we employ an omniscient perspective in our data generation process, where both seekers and supporters can access the complete information. While this approach is common in previous studies \citep{acl23findings-augesc, zheng2024self}, it doesn't fully reflect the real-world conversation dynamics. Future researchers may improve realism by simulating emotional support scenarios with distinct information states for each role.

\section*{Ethical Considerations}
Emotional support conversations (ESC) in LLM-generated dialogues requires careful ethical considerations.
We recognize the risks of using LLMs to generate emotional support responses, especially if the system misinterprets or misrepresents the user's persona, potentially leading to unintended emotional harm.
Users must clearly understand they are interacting with chatbots, not a human, to manage expectations and avoid misleading attachments.
Incorporating personas could make LLMs seem more human-like and more guiding, increasing the risk of user dependency on chatbots instead of seeking professional help. Therefore, implementing safeguards to guide users to human assistance in cases of severe distress is crucial.
While the LLMs demonstrate the ability to identify and utilize personas, they also inherit issues like societal biases, the risk of emotional manipulation, and dependency on LLM-generated support.
To mitigate these concerns, we emphasize that this research should only be considered for academic purposes and cannot be deployed in real-life emotional support scenarios without additional safeguards.
We are committed to improving ESC chatbots to minimize biases, enhance transparency, and support the development of more adaptive and ethically sound emotional support chatbots in the future.

\bibliography{acl_latex}

\appendix

\section{Prompts for Generating Persona Cards}
\label{sec:persona-card-prompt}
\begin{figure}
    \centering
    \includegraphics[width=1\linewidth]{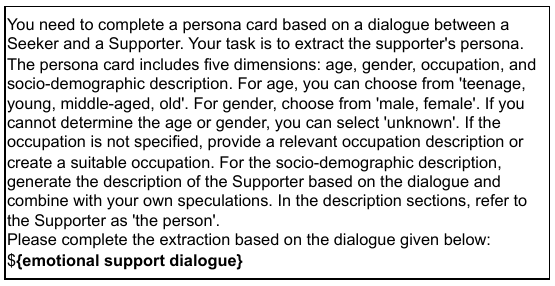}
    \caption{Prompt for extracting the basic persona from the dialogue.}
    \label{fig:prompt-basic-persona-dialogue}
\end{figure}

\begin{figure}
    \centering
    \includegraphics[width=0.98\linewidth]{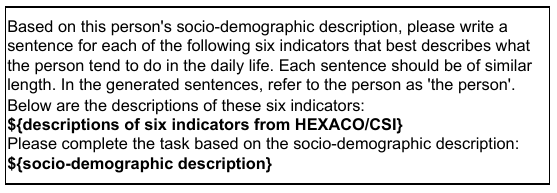}
    \caption{Prompt for producing the best describes on HEXACO/CSI indicators.}
    \label{fig:prompt-perdict-hexaco-csi-sentences}
    \vspace{-0.3cm}
\end{figure}

\begin{figure}
    \centering
    \includegraphics[width=0.98\linewidth]{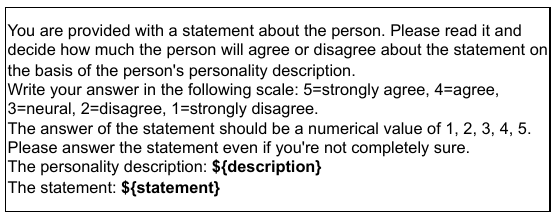}
    \caption{Prompt for answering HEXACO/CSI inventories based on persona.}
    \label{fig:prompt-questionnaires}
    \vspace{-0.3cm}
\end{figure}

\begin{figure}
    \centering
    \includegraphics[width=0.98\linewidth]{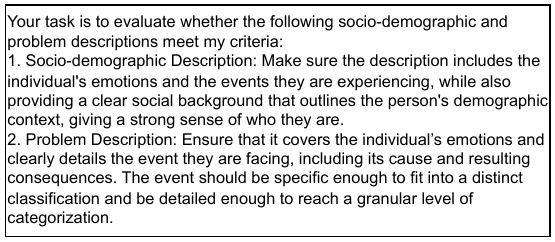}
    \caption{Prompt for filtering personas.}
    \label{fig:prompt-filter}
    \vspace{-0.3cm}
\end{figure}

In the section, it shows the prompts for generating persona cards, including basic persona and persona traits.
The prompt in Figure \ref{fig:prompt-basic-persona-dialogue} is used to extract basic persona information from dialogues, including age, gender, occupation and socio-demographic description.
Figure \ref{fig:prompt-perdict-hexaco-csi-sentences} presents the prompt used to determine the most suitable description of HEXACO or CSI indicators based on the extracted socio-demographic description.
Figure \ref{fig:prompt-questionnaires} shows the prompt used by LLMs to answer HEXACO and CSI inventories derived from the extracted personas.
Figure \ref{fig:prompt-filter} provides the prompt used to filter out unclear personas and those that do not provide identifiable identity information.

\section{Prompts for Measuring the Stability of Inferring Persona Traits}
\label{sec:prompt-stability}

\begin{figure}
    \centering
    \includegraphics[width=0.98\linewidth]{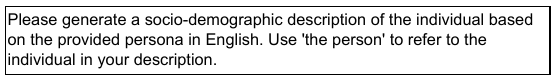}
    \caption{Prompt for extending personas from Persona Hub.}
    \label{fig:prompt-extend}
    \vspace{-0.3cm}
\end{figure}

\begin{figure}[t]
    \centering
    \includegraphics[width=0.98\linewidth]{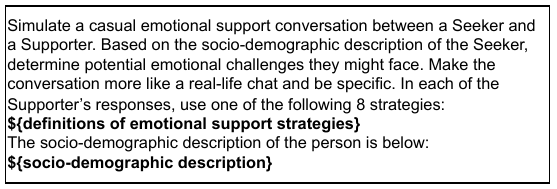}
    \caption{Prompt for generating emotional support dialogue based on the given persona.}
    \label{fig:prompt-extend-dialogue}
    \vspace{-0.3cm}
\end{figure}

In this section, we introduce the prompts used to measure whether LLMs can infer stable traits from persona in emotional support dialogues.
Figure \ref{fig:prompt-extend} shows the prompt used to extend the socio-demographic description of the persona.
Figure \ref{fig:prompt-extend-dialogue} displays the prompt to generate emotional support dialogues with strategies based on the given persona.
Appendix \ref{sec:persona-card-prompt} lists the prompts used to obtain HEXACO and CSI scores from personas.

\section{Correlations on Other Datasets}
\label{sec:other-correlations}
In this section, we introduce the correlations of HEXACO and CSI on the CAMS (Figure \ref{tab:gpt-CAMS}, \ref{tab:cams-seeker-claude} and \ref{tab:cams-seeker-llama}) and Dreaddit (Figure \ref{tab:gpt-Dreaddit}, \ref{tab:dreaddit-seeker-claude} and \ref{tab:dreaddit-seeker-llama}) dataset. The findings are same as in the Section \ref{sec:measure-traits}.

\begin{table}[t]
\begin{minipage}{0.48\textwidth}
    \centering
     \small
    \begin{tabular}{c|cccccc} \toprule
               & Expr. & Prec. & Verb. & Ques. & Emot. & Impr. \\\midrule
         Extr. & \cellcolor{mycolor_green}{\textbf{.66}} & .35 & -.27 & .52 & -.34 & -.02 \\
         Cons. & .43 & \cellcolor{mycolor_green}{\textbf{.40}} & -.34 & .36 & -.31 & -.01 \\
         Agre. & .40 & .24 & \cellcolor{mycolor_green}{\textbf{-.42}} & .30 & -.19 & .01 \\
         Open. & .54 & .36 & -.25 & \cellcolor{mycolor_green}{\textbf{.53}} & -.16 & .02 \\
         Emot. & -.25 & -.22 & .01 & -.20 & \cellcolor{mycolor_green}{\textbf{.59}} & -.02 \\
         Hone. & -.18 & -.05 & -.19 & -.14 & .13 & \cellcolor{mycolor_green}{\textbf{-.15}} \\\bottomrule
    \end{tabular}
    \caption{Correlations between CSI and HEXACO of CAMS personas measured by \textit{gpt-40-mini}.}
    \label{tab:gpt-CAMS}
 \vspace{3mm}
\end{minipage}
\begin{minipage}{0.48\textwidth}
    \centering
     \small
        \begin{tabular}{c|cccccc} \toprule
                   & Expr. & Prec. & Verb. & Ques. & Emot. & Impr. \\\midrule
             Extr. & \cellcolor{mycolor_green}{\textbf{.73}} & .09 & -.30 & \cellcolor{mycolor_red}{\textbf{.54}} & .20 & -.22 \\
             Cons. & .29 & \cellcolor{mycolor_green}{\textbf{.47}} & -.45 & .21 & -.08 & -.29 \\
             Agre. & .26 & .19 & \cellcolor{mycolor_green}{\textbf{-.61}} & .10 & .09 & -.32 \\
             Open. & .44 & .17 & -.39 & \cellcolor{mycolor_green}{.42} & .19 & -.26 \\
             Emot. & -.24 & -.24 & .00 & -.19 & \cellcolor{mycolor_green}{\textbf{.30}} & .02 \\
             Hone. & .15 & .31 & -.55 & .08 & .06 & \cellcolor{mycolor_green}{\textbf{-.39}} \\\bottomrule
            \end{tabular}
            \caption{Correlations between CSI and HEXACO from CAMS measured by \textit{Claude-3.5-haiku}.}
            \label{tab:cams-seeker-claude}
 \vspace{3mm}
\end{minipage}
\begin{minipage}{0.48\textwidth}
    \centering
     \small
        \begin{tabular}{c|cccccc} \toprule
                   & Expr. & Prec. & Verb. & Ques. & Emot. & Impr. \\\midrule
             Extr. & \cellcolor{mycolor_green}{\textbf{.44}} & .17 & -.25 & .27 & -.35 & \cellcolor{mycolor_red}{-.17} \\
             Cons. & .27 & \cellcolor{mycolor_green}{\textbf{.44}} & \cellcolor{mycolor_red}{-.34} & .10 & -.36 & \cellcolor{mycolor_red}{-.18} \\
             Agre. & .12 & .06 & \cellcolor{mycolor_green}{-.32} & .09 & -.02 & -.11 \\
             Open. & .25 & -.03 & -.16 & \cellcolor{mycolor_green}{\textbf{.47}} & .06 & -.16 \\
             Emot. & .26 & -.29 & .15 & -.11 & \cellcolor{mycolor_green}{\textbf{.50}} & .15 \\
             Hone. & -.06 & .03 & .18 & -.09 & .05 & -.01 \\\bottomrule
            \end{tabular}
            \caption{Correlations between CSI and HEXACO from CAMS measured by \textit{LLaMA-3.1-8B-Instruct}.}
            \label{tab:cams-seeker-llama}
    \end{minipage}
\end{table}

\begin{table}[t]
    \begin{minipage}{0.48\textwidth}
        
    \centering
    \small
    \begin{tabular}{c|cccccc} \toprule
               & Expr. & Prec. & Verb. & Ques. & Emot. & Impr. \\\midrule
         Extr. & \cellcolor{mycolor_green}{\textbf{.59}} & .26 & -.25 & .42 & -.55 & -.02 \\
         Cons. & .34 & \cellcolor{mycolor_green}{\textbf{.36}} & -.24 & .22 & -.40 & -.05 \\
         Agre. & .22 & .20 & \cellcolor{mycolor_green}{\textbf{-.43}} & .12 & -.22 & -.12 \\
         Open. & .45 & .28 & -.20 & \cellcolor{mycolor_green}{\textbf{.49}} & -.21 & .05 \\
         Emot. & -.39 & -.17 & .05 & -.27 & \cellcolor{mycolor_green}{\textbf{.64}} & .00 \\
         Hone. & -.31 & .00 & -.20 & -.27 & .15 & \cellcolor{mycolor_green}{\textbf{-.26}} \\\bottomrule
    \end{tabular}
    \caption{Correlations between CSI and HEXACO of Dreaddit personas measured by \textit{gpt-40-mini}.}
    \label{tab:gpt-Dreaddit}
 \vspace{3mm}
\end{minipage}
\begin{minipage}{0.48\textwidth}
    \centering
     \small
        \begin{tabular}{c|cccccc} \toprule
                   & Expr. & Prec. & Verb. & Ques. & Emot. & Impr. \\\midrule
             Extr. & \cellcolor{mycolor_green}{\textbf{.67}} & -.01 & -.17 & \cellcolor{mycolor_red}{\textbf{.52}} & .13 & -.11 \\
             Cons. & .11 & \cellcolor{mycolor_green}{\textbf{.46}} & -.28 & .07 & -.16 & -.28 \\
             Agre. & .12 & .14 & \cellcolor{mycolor_green}{\textbf{-.59}} & -.01 & .02 & -.37 \\
             Open. & .38 & .10 & -.29 & \cellcolor{mycolor_green}{.44} & .22 & -.12 \\
             Emot. & -.37 & -.17 & -.02 & -.30 & \cellcolor{mycolor_green}{\textbf{.26}} & .03 \\
             Hone. & -.01 & .34 & -.49 & -.08 & -.01 & \cellcolor{mycolor_green}{\textbf{-.42}} \\\bottomrule
            \end{tabular}
            \caption{Correlations between CSI and HEXACO from Dreaddit measured by \textit{Claude-3.5-haiku}.}
            \label{tab:dreaddit-seeker-claude}
 \vspace{3mm}
\end{minipage}
\begin{minipage}{0.48\textwidth}
    \centering
     \small
        \begin{tabular}{c|cccccc} \toprule
                   & Expr. & Prec. & Verb. & Ques. & Emot. & Impr. \\\midrule
             Extr. & \cellcolor{mycolor_green}{\textbf{.41}} & .26 & -.30 & .28 & -.31 & \cellcolor{mycolor_red}{\textbf{-.29}} \\
             Cons. & .19 & \cellcolor{mycolor_green}{\textbf{.55}} & \cellcolor{mycolor_red}{\textbf{-.45}} & .17 & -.35 & -.27 \\
             Agre. & .20 & .06 & \cellcolor{mycolor_green}{-.33} & .07 & -.06 & -.21 \\
             Open. & .24 & .10 & -.22 & \cellcolor{mycolor_green}{\textbf{.37}} & -.04 & -.27 \\
             Emot. & -.10 & -.24 & .13 & -.13 & \cellcolor{mycolor_green}{\textbf{.43}} & .09 \\
             Hone. & -.06 & .01 & -.16 & -.03 & .03 & -.18 \\\bottomrule
            \end{tabular}
            \caption{Correlations between CSI and HEXACO from Dreaddit measured by \textit{LLaMA-3.1-8B-Instruct}.}
            \label{tab:dreaddit-seeker-llama}
    \end{minipage}
\end{table}

\section{Prompts for Synthesizing Dialogues w/o Persona Traits}
\label{sec:prompt-synthesize-wo-traits}
\begin{figure}[t]
    \centering
    \includegraphics[width=0.98\linewidth]{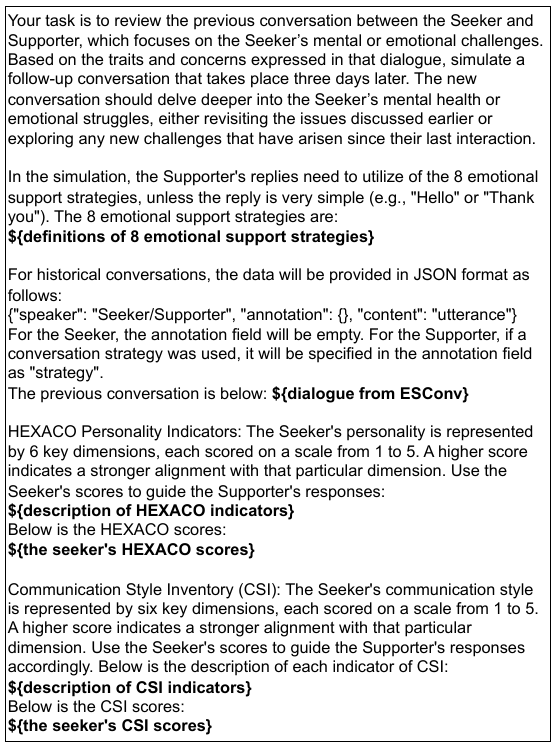}
    \caption{Prompt for synthesizing dialogues with persona traits.}
    \label{fig:prompt-perdict-synthesize-w-traits}
    \vspace{-0.3cm}
\end{figure}

\begin{figure}[t]
    \centering
    \includegraphics[width=0.98\linewidth]{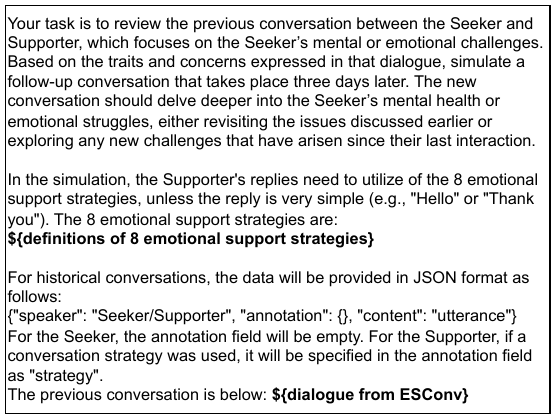}
    \caption{Prompt for synthesizing dialogues without persona traits.}
    \label{fig:prompt-perdict-synthesize-wo-traits}
    \vspace{-0.3cm}
\end{figure}

In this part, we give prompts for synthesizing emotional support dialogues with and without persona traits continuing writing ESConv dialogues.
Figure \ref{fig:prompt-perdict-synthesize-w-traits} illustrates the prompt to generate dialogues with personas, while the prompt shown in Figure \ref{fig:prompt-perdict-synthesize-wo-traits} is utilized to generate dialogues without personas.

\section{Results of Persoan Consistency}
\label{sec:persona-consistency}
\begin{figure}[t]
    \centering
    \begin{minipage}{0.46\textwidth}
        \centering
        \includegraphics[width=\textwidth]{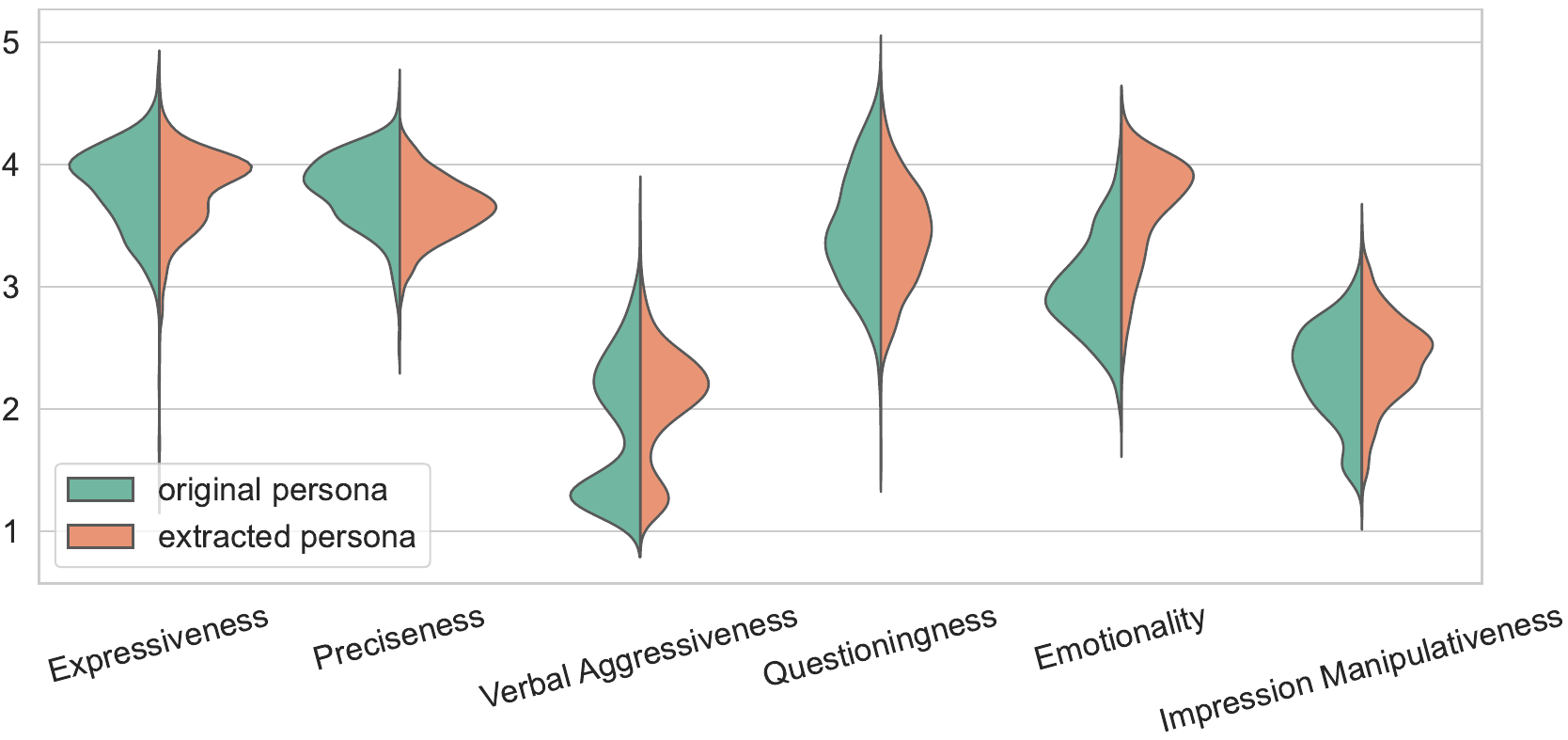}
        \caption{Comparison of CSI scores between the original persona and the one extracted from the dialogue generated by \textit{gpt-4o-mini}.}
        \label{fig:csi-violin}
    \end{minipage}
    \begin{minipage}{0.46\textwidth}
        \centering
        \includegraphics[width=\textwidth]{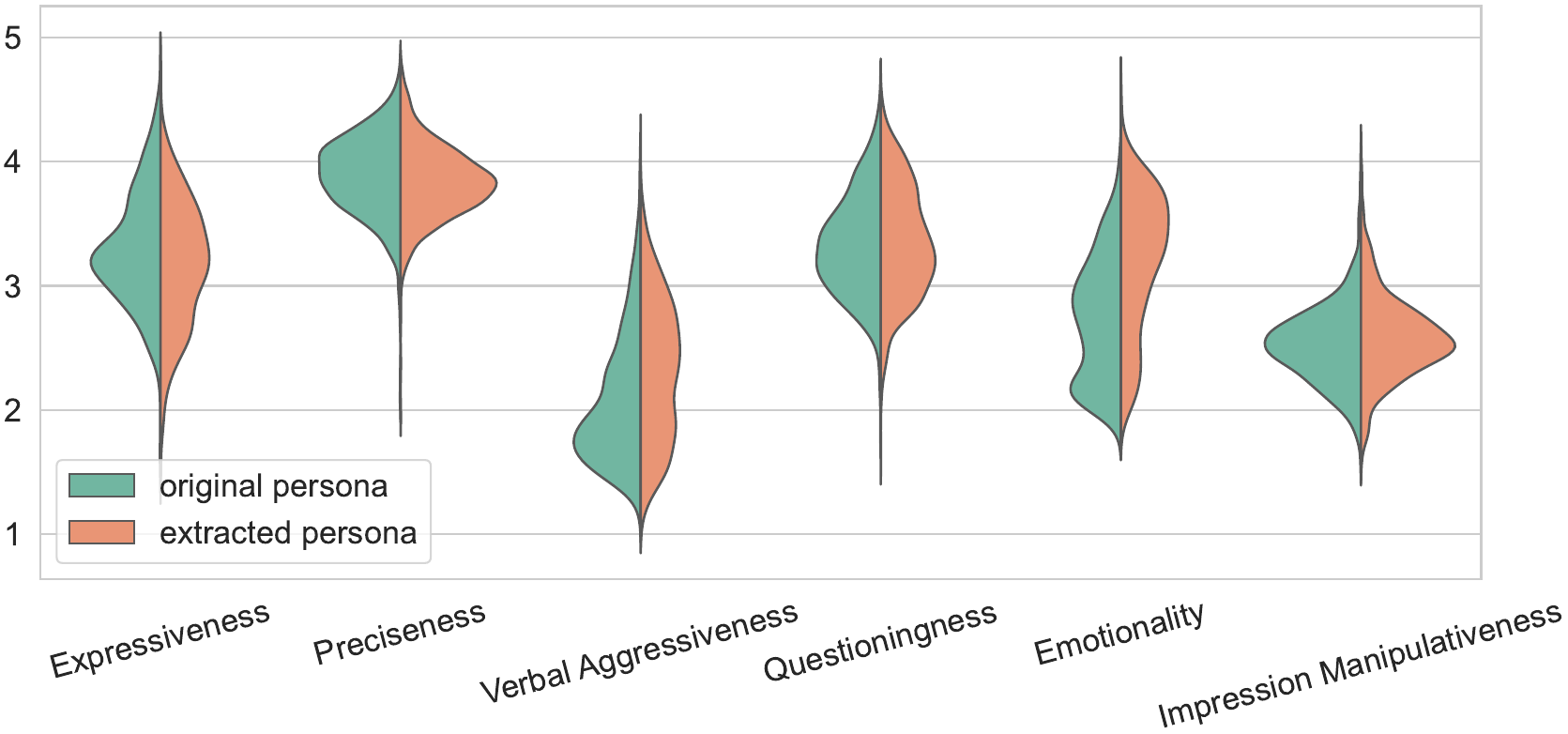}
        \caption{Comparison of CSI scores between the original persona and the one extracted from the dialogue generated by \textit{claude-3.5-haiku}.}
        \label{fig:csi-violin-claude}
    \end{minipage}
    \begin{minipage}{0.46\textwidth}
        \centering
        \includegraphics[width=\textwidth]{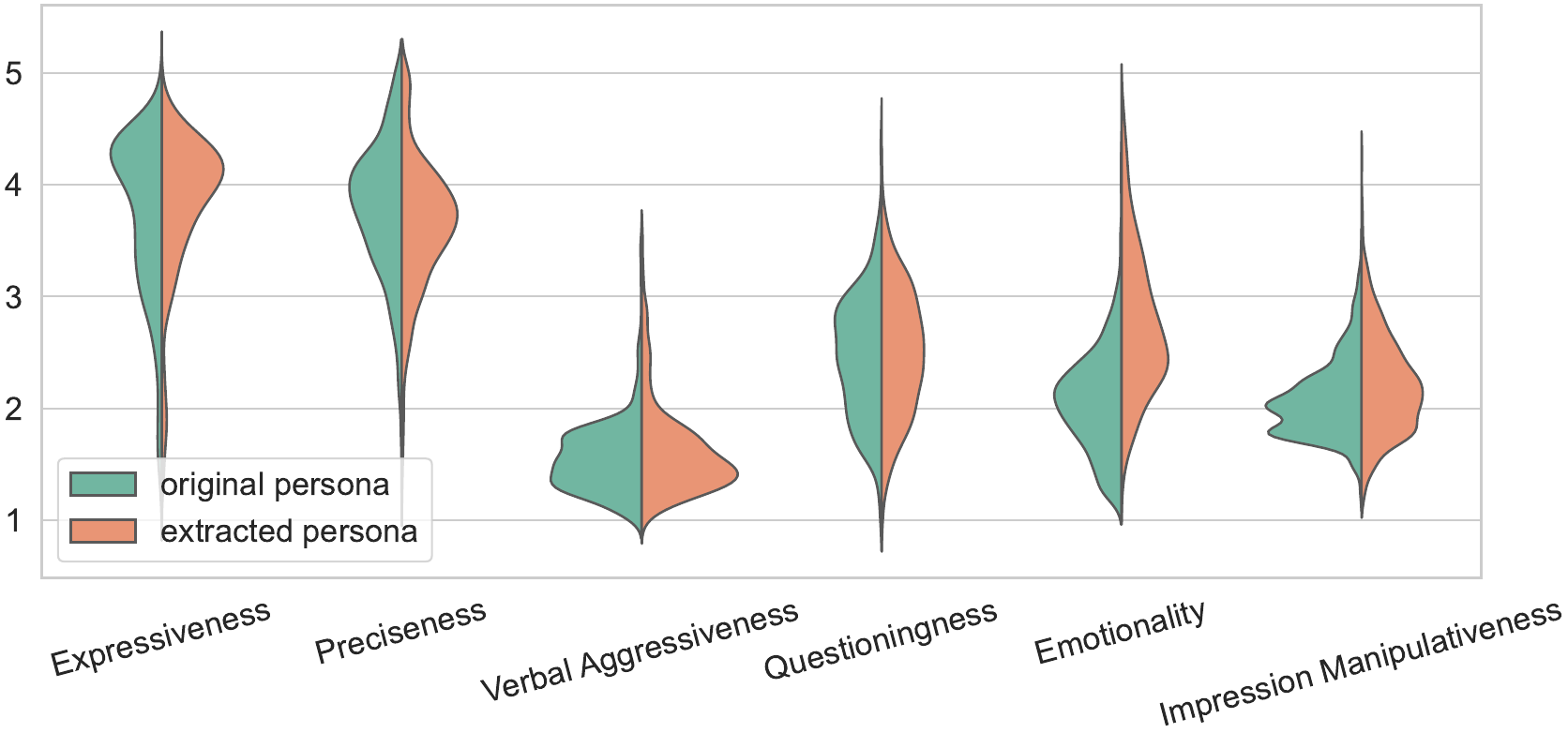}
        \caption{Comparison of CSI scores between the original persona and the one extracted from the dialogue generated by \textit{LLaMA-3.1-8B-Instruct}.}
        \label{fig:csi-violin-llama}
    \end{minipage}
    \vspace{-0.3cm}
\end{figure}

Table \ref{fig:hexaco-violin-claude} and \ref{fig:csi-violin-claude} shows the results of measuring the communication style consistency after synthesizing emotional support dialogues by \textit{GPT-40-mini, Claude-3.5-Haiku, LLaMA-3.1-8B-Instruct}. The results are comparable to those of other models discussed in Section \ref{sec:rq2-results}.

\begin{figure}[t]
    \centering
    \includegraphics[width=0.98\linewidth]{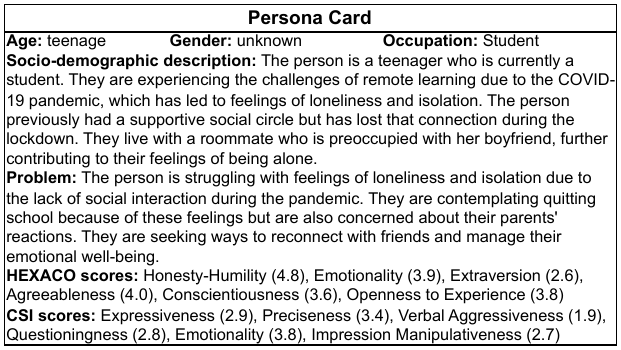}
    \caption{The persona card of the case study.}
    \label{fig:persona-card-case-study}
    \vspace{-0.3cm}
\end{figure}

\begin{figure}[t]
    \centering
    \includegraphics[width=0.98\linewidth]{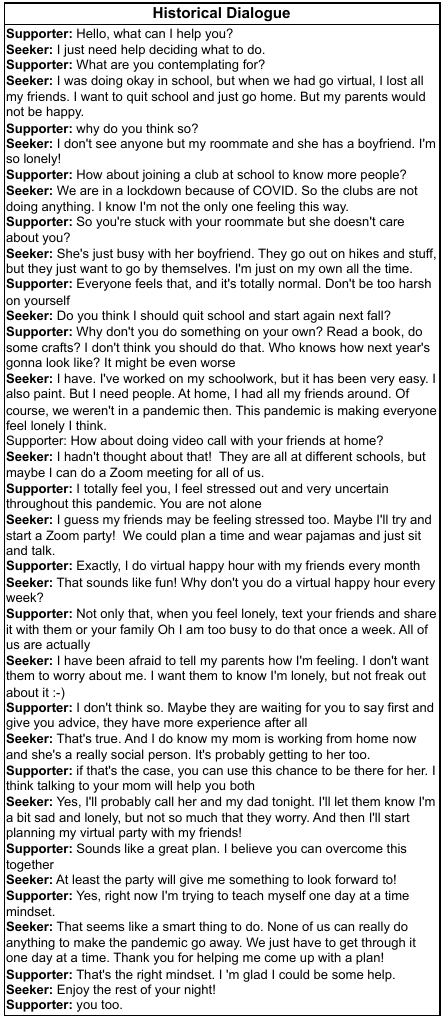}
    \caption{The historical dialogue of the case study.}
    \label{fig:historical-dialogue}
    \vspace{-0.3cm}
\end{figure}

\section{Persona Card and Dialogue History of Case Study}
\label{sec:case-study}

Figure \ref{fig:persona-card-case-study} shows the persona card injected into the generation, while Figure \ref{fig:historical-dialogue} shows the historical dialogue of the case from ESConv.

\section{Definitions of Emotional Support Strategies}
\label{sec:definitions-strategies}
In our experiments, we use the emotional support strategies used in ESConv dataset \citep{esconv}. The definitions of emotional support strategies are shown below:\\
\textbf{Question:} Asking open-ended or specific questions related to the problem to help the seeker articulate their issues and provide clarity.\\
\textbf{Restatement or Paraphrasing:} Concisely rephrasing the seeker’s statements to help them better understand their situation.\\
\textbf{Reflection of feelings:} Expressing and clarifying the seeker’s emotions to acknowledge their feelings.\\
\textbf{Self-disclosure:} Sharing similar experiences or emotions to build empathy and connection.\\
\textbf{Affirmation and Reassurance:} Affirm the seeker’s strengths, motivation, and capabilities and provide reassurance and encouragement.\\
\textbf{Providing Suggestions:} Offering possible ways forward while respecting the seeker’s autonomy.\\
\textbf{Information:} Provide useful information to the seeker.\\
\textbf{Others:} Exchange pleasantries and use strategies beyond the defined categories.

\section{Dialogue Generation Statistics}

\begin{table}
    \centering
    \small
\begin{tabular}{l|cccccc} \toprule
           & \textbf{w/ persona} & \textbf{w/o personas} \\\midrule
     Total Words & 218,433 & 232,674 \\
     Total Turns & 10,398 & 12,666 \\
     Avg Words (Total) & 21.01 & 18.37 \\
     Seeker Words & 91,590 & 94,286 \\
     Seeker Turns & 5,199 & 6,323 \\
     Avg Words (Seeker) & 17.62 & 14.91 \\
     Supporter Words & 126,843 & 138,388 \\
     Supporter Turns & 5,199 & 6,343 \\
     Avg Words (Supporter) & 24.40 & 21.82 \\\bottomrule
\end{tabular}
\caption{Statistics of generated dialogues.}
\label{tab:statistics-generated-dialogues}
\end{table}

Table \ref{tab:statistics-generated-dialogues} shows the statistics of dialogues generated w/ and w/o personas. We can observe dialogues generated with personas have fewer turns but are substantially longer on average per turn for both seeker and supporter. This aligns with our qualitative findings (Case study, Figure \ref{fig:case-short}): the persona-guided supporter asks more targeted questions, leading to more substantive replies from the seeker and a more efficient, in-depth conversation.

\end{document}